\documentclass[runningheads]{llncs}
\usepackage{comment}
\usepackage{graphicx, amsmath, amssymb, caption, subcaption, multirow, overpic, textpos}
\usepackage[table]{xcolor}
\usepackage[british, english, american]{babel}
\usepackage[pagebackref=false, breaklinks=true, letterpaper=true, colorlinks,
            citecolor=citecolor, linkcolor=linkcolor, bookmarks=false]{hyperref}

\definecolor{citecolor}{HTML}{0071BC}
\definecolor{linkcolor}{HTML}{ED1C24}

\definecolor{darkpastelgreen}{rgb}{0.01, 0.75, 0.24}
\definecolor{darkgray}{rgb}{0.66, 0.66, 0.66}
\definecolor{bittersweet}{rgb}{1.0, 0.44, 0.37}
\definecolor{bleudefrance}{rgb}{0.19, 0.55, 0.91}
\definecolor{ao(english)}{rgb}{0.0, 0.5, 0.0}
\newlength\savewidth\newcommand\shline{\noalign{\global\savewidth\arrayrulewidth
  \global\arrayrulewidth 1pt}\hline\noalign{\global\arrayrulewidth\savewidth}}
\newcommand{\tablestyle}[2]{\setlength{\tabcolsep}{#1}\renewcommand{\arraystretch}{#2}\centering\footnotesize}

  \newcolumntype{x}[1]{>{\centering\arraybackslash}p{#1pt}}
\newcolumntype{y}[1]{>{\raggedright\arraybackslash}p{#1pt}}
\newcolumntype{z}[1]{>{\raggedleft\arraybackslash}p{#1pt}}
\usepackage[capitalize]{cleveref}

\newcommand{\app}{\raise.17ex\hbox{$\scriptstyle\sim$}}

\definecolor{deemph}{gray}{0.7}
\newcommand{\gc}[1]{\textcolor{gray}{#1}}
\definecolor{baselinecolor}{gray}{.9}
\newcommand{\baseline}[1]{\cellcolor{baselinecolor}{#1}}

\crefname{section}{Sec.}{Secs.}
\Crefname{section}{Section}{Sections}
\Crefname{table}{Table}{Tables}
\crefname{table}{Tab.}{Tabs.}
\def\onedot{.}
\def\eg{\emph{e.g}\onedot} 
\def\ie{\emph{i.e}\onedot}

\begin{document}
\pagestyle{headings}
\mainmatter

\title{ConCL: Concept Contrastive Learning for Dense Prediction Pre-training in Pathology Images} 

\titlerunning{ConCL: Concept Contrastive Learning for Dense Pre-training in Pathology}

\author{Jiawei Yang\inst{1,2,}\thanks{Work done during an internship at Tencent AI Lab.} \and Hanbo Chen\inst{1,}\thanks{Corresponding authors} \and Yuan Liang\inst{2} \and Junzhou Huang \inst{3} \and Lei He\inst{2} \and Jianhua Yao \inst{1,\star\star}}

\authorrunning{J. Yang et al.}

\institute{Tencent AI Lab \and University of California, Los Angeles \\ \and University of Texas at Arlington \\ \email{jiawei118@ucla.edu}}
%%%%%%%%%%%%%%%%%%%%%%%%%%%%%%%%%%%%%%%%%%%%%%%%%%%%%%%%%%%%%%%%%%%%%%%%%%%%%%%%%%%%%%%%%%%%%%%%%%%
\maketitle
\begin{abstract}
Detecting and segmenting objects within whole slide images is essential in computational pathology workflow. 
Self-supervised learning (SSL) is appealing to such annotation-heavy tasks. Despite the extensive benchmarks in natural images for \textit{dense} tasks, such studies are, unfortunately, absent in current works for pathology. Our paper intends to narrow this gap. We first benchmark representative SSL methods for dense prediction tasks in pathology images. Then, we propose \textbf{con}cept \textbf{c}ontrastive \textbf{l}earning (ConCL), an SSL framework for dense pre-training. We explore how ConCL performs with concepts provided by different sources and end up with proposing a simple dependency-free concept generating method that does not rely on external segmentation algorithms or saliency detection models. Extensive experiments demonstrate the superiority of ConCL over previous state-of-the-art SSL methods across different settings. Along our exploration, we distill several important and intriguing components contributing to the success of dense pre-training for pathology images. We hope this work could provide useful data points and encourage the community to conduct ConCL pre-training for problems of interest. 
Code is available at \url{https://github.com/TencentAILabHealthcare/ConCL}.

\end{abstract}
%%%%%%%%%%%%%%%%%%%%%%%%%%%%%%%%%%%%%%%%%%%%%%%%%%%%%%%%%%%%%%%%%%%%%%%%%%%%%%%%%%%%%%%%%%%%%%%%%%%
\section{Introduction}
\label{sec:intro}

%##################################################################################################
\begin{figure}[t]\centering
\includegraphics[width=\linewidth]{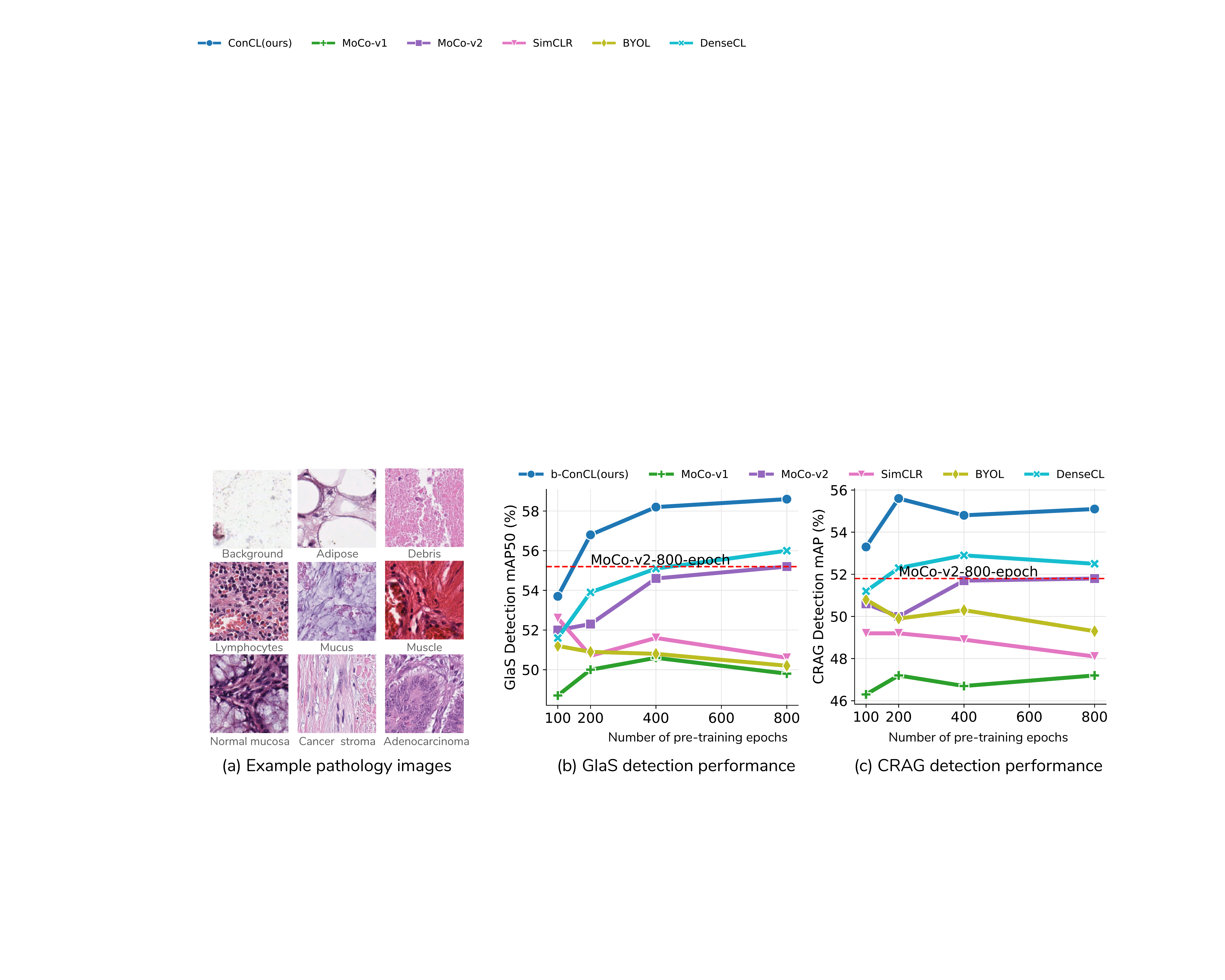}
\caption{(a) Example pathology images with tissue class names. (b,c) Comparisons of pre-trained models by fine-tuning on GlaS dataset \cite{sirinukunwattana2017gland} and CRAG dataset \cite{graham2019mild}. All models are fine-tuned with the same detector, \ie, Mask-RCNN architecture \cite{he2017mask}, ResNet-18 backbone \cite{he2016deep}, and FPN head\cite{lin2017feature} under $1\times$ fine-tuning schedule. Results are averaged over 5 independent runs.}
\label{fig:longer_pretraining1}
\end{figure}
%##################################################################################################

Computational pathology is an emerging area in modern healthcare. More whole slide images (WSIs) are now analyzed by deep learning (DL) models \cite{srinidhi2021deep}. To alleviate the heavy annotation burden required by DL models, reusing weights from pre-trained models has become a common practice. Besides transferring from fully-supervised models, recent attention has been attracted to self-supervised learning (SSL) methods \cite{he2020momentum,chen2020simple,grill2020bootstrap}. They are annotation-free but can achieve comparable or even better performance when transferring.

The breakthrough of SSL methods starts with contrastive learning \cite{hadsell2006dimensionality,wu2018unsupervised,chen2020simple,he2020momentum,chen2020improved}, where the most popular task is instance discrimination \cite{wu2018unsupervised}. It requires a model to discriminate among individual instances, \ie, image-level representations.To achieve that, it first defines some positive pairs and negative pairs. It then optimizes a model to maximize the representation similarity between positive pairs and minimize it between negative pairs. Later, more SSL methods based on cross-view prediction are proposed, \eg, \cite{caron2020unsupervised,grill2020bootstrap,chen2020simple,zbontar2021barlow}. However, these methods are optimized for image-level representations and might be sub-optimal for dense prediction tasks such as object detection and instance segmentation. This motivates works for detection-friendly pre-training methods, \eg, DenseCL \cite{wang2021dense}, InsLoc \cite{yang2021instance}, \textit{Self}-EMD \cite{liu2020self}, SCRL \cite{roh2021spatially}, and more \cite{henaff2021efficient,van2021unsupervised,xie2021detco,xie2021propagate}. Despite many interests raised in the natural image domain for dense prediction problems, such studies, which are of important clinical and practical values, are \textit{absent} in the pathology image domain. Our research is intended to bridge the gap between SSL in natural images and pathology images for dense prediction tasks, as well as to distill the key components to the success of dense pre-training in the pathology data. 

To that end, we start by presenting a general \textbf{Con}cept \textbf{C}ontrastive \textbf{L}earning (ConCL) framework. Rather than contrasting image-level representations \cite{wu2018unsupervised,chen2020simple,he2020momentum}, it contrasts ``concepts'' that mark different local (semantic) regions. ConCL is an abstraction of dense contrasting frameworks and thus, as we show later, resembles most concurrent related works. We first benchmark current leading image-level SSL methods and a grid-level dense SSL method (\ie, DenseCL \cite{wang2021dense}) in two public datasets. We observe a considerable performance gap between DenseCL \cite{wang2021dense} and the others. These gaps indicate the importance of contrasting densely (grid-level) than roughly (image-level). Then, directed by the performance differences and the characteristics of pathology images, we gradually develop and improve ConCL via a series of explorations. Specifically, we explore: 1) \textit{what makes the success of dense prediction pre-training?} 2) \textit{what kind of concepts are good for pathology images?} The nature of having rich low-level patterns in pathology images (see \cref{fig:longer_pretraining1}-(a)) gives some surprising and intriguing results, \eg, a randomly initialized model can group meaningful concepts and help dense pre-training. Along the exploration, we distill several key components contributing to the transferring performance for dense tasks. At the end of exploration, the presented ConCL can surpass various state-of-the-art SSL methods by solid and consistent margins across different downstream datasets, detector architectures, fine-tuning schedules, and pre-training epochs. For example, as shown in \Cref{fig:longer_pretraining1}-(b), the 200-epoch pre-trained ConCL wins all the other methods but with $4\times$ to $8\times$ fewer epochs. To summarize, this paper makes the following contributions:
\begin{itemize}
    \item[$\bullet$] It makes one of the earliest attempts to systematically study and benchmark self-supervised learning methods for dense prediction problems in pathology images, which are of high practical and clinical interest but, unfortunately, \textit{absent} in existing works. We hope this work could narrow the gap between studies in natural images and pathology images.
    \item[$\bullet$] It presents ConCL, an SSL framework for dense pre-training. We show how ConCL performs with concepts provided by different sources and find that a randomly initialized model could learn semantic concepts and improve itself without expert-annotation or external algorithms while achieving competitive, if not the best, results.
    \item[$\bullet$] It shows how important the \textit{dense} pre-training is in pathology images for dense tasks and provides some intriguing observations that could contribute to other applications such as few-shot and semi-supervised segmentation and detection, or more, in pathology image analysis or beyond. 
\end{itemize}
We hope this work could provide useful data points and encourage the community to conduct ConCL pre-training for problems of interest.

%%%%%%%%%%%%%%%%%%%%%%%%%%%%%%%%%%%%%%%%%%%%%%%%%%%%%%%%%%%%%%%%%%%%%%%%%%%%%%%%%%%%%%%%%%%%%%%%%%%
\section{Related work}
\subsubsection{Contrastive learning.} The success of deep learning is mainly attributed to mining a large amount of data. When limited data is provided for specific tasks, an alternative is to transfer knowledge by re-using pre-trained models \cite{girshick2014rich,he2019rethinking}. SSL methods learn good pre-trained models from label-free pretext tasks, \eg, colorization \cite{zhang2016colorful,zhang2017split}, denoising \cite{vincent2008extracting}, and thus attract much attention. Recently, contrastive learning \cite{he2020momentum,chen2020improved,chen2020simple,oord2018representation,wu2018unsupervised,caron2020unsupervised}, a typical branch of SSL, has made significant progress in many fields, where instance discrimination \cite{hadsell2006dimensionality,wu2018unsupervised,he2020momentum,chen2020improved,chen2020simple} serves as a pretext task. It requires a model to discriminate among individual instances, \ie, image-level representations \cite{wu2018unsupervised}. MoCo \cite{he2020momentum,chen2020improved} and SimCLR\cite{chen2020simple} are two representatives. Specifically, they generate two views of the same image via random data augmentations (\eg, color jittering, random cropping) and mark them as a positive pair. Then, views from other different images are marked as negative instances or pairs. After that, they learn embeddings by maximizing the similarity between the representations of positive pairs while minimizing it between the representations of negative pairs. Later methods combine contrasting with clustering, \eg, SwAV\cite{caron2020unsupervised} proposes to contrast views' cluster assignments, and PCL \cite{li2020prototypical} contrasts instances with cluster prototypes. 

\subsubsection{Dense prediction pre-training.} Despite their success in transferring to classification tasks, good image-level representations do not necessarily result in better performance in dense prediction tasks. Therefore, recent efforts have been made for dense prediction pre-training. Related works are mostly concurrent \cite{wang2021dense,yang2021instance,roh2021spatially,van2021unsupervised,xie2021detco,xie2021propagate,liu2020self,henaff2021efficient}. Among them, DenseCL \cite{wang2021dense} learns the correspondence among pixels of a positive pair and optimizes a pairwise contrastive loss at a pixel level, yielding a dense contrasting behavior. \textit{Self}-EMD \cite{liu2020self} does dense predicting in a non-contrastive manner as in BYOL \cite{grill2020bootstrap}, \ie, predicting a grid-level feature vector from one view when given its counterpart from another (positive) view. SCRL \cite{roh2021spatially} argues the importance of spatially consistent representations, so it maximizes the similarity of box region features in the intersected area. The most relevant works concurrent to ours are \cite{henaff2021efficient,van2021unsupervised}. They also optimize contrastive loss over mask-averaged representations. Those masks are generated by external algorithms that are successful for natural images, \eg, Felzenszwalb-Huttenlocher algorithm \cite{felzenszwalb2004efficient}, or models, \eg, MCG\cite{arbelaez2014multiscale}, BASNet\cite{qin2019basnet}, and DeepUSPS\cite{nguyen2019deepusps}. However, the success of such mask generators is unfortunately unverified in pathology images. In this paper, we provide some of their empirical results. Their different performances yield the disparity between natural and pathology images, from which we are motivated to propose a dependency-free concept mask generator. It directly bootstraps the structural concepts inherent in pathology images, learns from \textit{scratch}, and has better potential.

\subsubsection{SSL in pathology images.} Studying SSL methods in pathology images is still at an early stage. In addition to studies on natural images, SimCLR \cite{chen2020simple} is also studied and benchmarked for classification, regression, and segmentation tasks in pathology images \cite{ciga2020self}. Some domain-specific self-supervised pretext tasks, \eg, magnification prediction, JigMag prediction, and hematoxylin channel prediction, are proposed and studied \cite{koohbanani2021self}. However, despite interest raised in natural images for dense problems, existing works have not studied, to our knowledge, detection/segmentation-friendly SSL methods in pathology images. Our work aims to bridge this gap and provide our exploration roadmap toward better dense prediction performance for pathology images.

%%%%%%%%%%%%%%%%%%%%%%%%%%%%%%%%%%%%%%%%%%%%%%%%%%%%%%%%%%%%%%%%%%%%%%%%%%%%%%%%%%%%%%%%%%%%%%%%%%%
\section{Method}
\label{sec:method}

We start by briefly reviewing MoCo \cite{he2020momentum,chen2020improved} and use it as a running example to describe instance contrastive learning \cite{hadsell2006dimensionality,wu2018unsupervised}. Then we derive our motivation for ConCL and describe its details.

\subsection{Preliminary: Instance Contrastive Learning}\label{sec:prelim}

MoCo\cite{he2020momentum} abstracts the instance discrimination task as a dictionary look-up problem. Specifically, for each encoded query $q$, there is a set of encoded keys $\{k_0, k_1, k_2, ...\}$ in a dictionary. The instance discrimination task is to pull closer $q$ and its matched positive key $k_+$ in the dictionary while spreading $q$ away from all other negative keys $k_-$. When using the dot-product as similarity measurement, a form of contrastive loss function based on InfoNCE\cite{oord2018representation} becomes:
\begin{equation}
\label{eq:moco_loss}
L_q = -\log\frac{\exp(q\cdot k_+ / \tau)}{\exp(q\cdot k_+ / \tau) + \sum_{k_{-}}\exp(q\cdot k_{-}/\tau)}
\end{equation}
where $\tau$ is a temperature hyper-parameter \cite{wu2018unsupervised}. Queries $q$ and keys $k$ are computed by a query encoder and a key encoder, respectively \cite{he2020momentum,chen2020improved}. Formally, $q=h(\verb|GAP|(f_5(x_q)))$, where $h$ is a MLP projection head as per \cite{chen2020simple}; $\verb|GAP|(\cdot)$ denotes global-average-pooling, and $f_5(x)$ represents the outputs from the stage-5 of a ResNet \cite{he2016deep}. Keys $k$ are computed similarly using the key encoder. In MoCo \cite{he2020momentum}, the negative keys are stored in a queue to avoid using large batches \cite{chen2020simple}.

%##################################################################################################
\begin{figure}[t]\centering
\includegraphics[width=\linewidth]{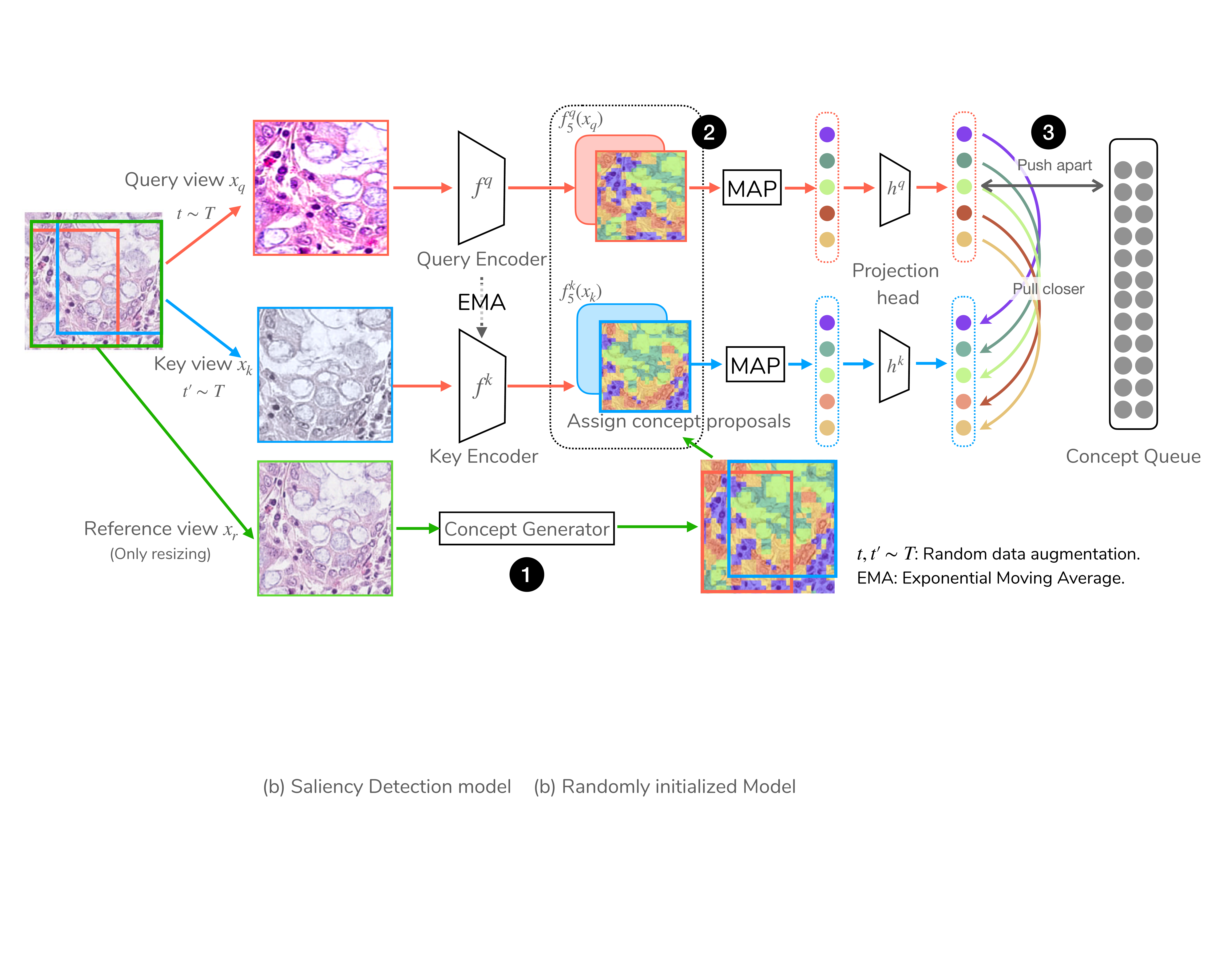}
\caption{\textbf{ConCL overview.} ConCL has three steps: (1) Given a query view {\color{bittersweet} $x_q$} and a key view {\color{bleudefrance} $x_k$}, their union region is cropped as a reference view {\color{ao(english)} $x_r$}. ConCL obtains concept proposals by processing {\color{ao(english)} $x_r$} with a ``concept generator.'' (2) For the shared concepts, ConCL computes their representations via masked average pooling (MAP). (3) ConCL optimizes concept contrastive loss (\cref{eq:concept_loss}), and enqueues the concept prototypes from the key encoder to the concept queue.}
\label{fig:overview}
\end{figure}
%##################################################################################################

\subsection{Concept Contrastive Learning}

Instance contrastive methods \cite{chen2020simple,he2020momentum,wu2018unsupervised} do well in discriminating among image-level instances, but dense prediction tasks usually require discriminating among local details, \eg, object instances or object parts. We abstract such local details, or say, fine-grained semantics as ``concepts.'' A concept does not necessarily represent an object. Instead, any sub-region in an image could be a concept since it contains certain different semantics.  From the perspective of dense prediction, it is desirable to build concept-sensitive representations. For example, one WSI patch usually contains multiple small objects, \eg, nucleus, glands, and multiple texture-like tissues, \eg, mucus \cite{srinidhi2021deep,kather_jakob_nikolas_2018_1214456}. To successfully detect and segment objects in such images, models need to learn more information from local details. To this end, we propose a simple but effective framework --- \textit{\textbf{Con}cept \textbf{C}ontrastive \textbf{L}earning} (ConCL). \Cref{fig:overview} shows its overview, which we elaborate on below.

\subsubsection{Concept discrimination.} We first define a pretext task named concept discrimination. Similar to instance discrimination \cite{wu2018unsupervised,hadsell2006dimensionality}, concept discrimination requires a model to discriminate among the representations of the same but augmented concepts and the representations of different concepts. We formulate concept discrimination by extending the instance-level queries and keys to concept-level. Specifically, given an encoded query concept $q^c$ and a set of encoded key concepts $\{k^c_0, k^c_1, k^c_2, ...\}$, we derive concept contrastive loss as:
\begin{equation}
\label{eq:concept_loss}
	L_c = -\log\frac{\exp(q^c\cdot k^c_+ / \tau)}{\exp(q^c\cdot k^c_+ / \tau) + \sum_{k^c_{-}}\exp(q^c\cdot k^c_{-}/\tau)}
\end{equation}
where $\tau$ is the same temperature parameter and $k_-^c$ are keys in the concept queue --- the queue to store concept representations. This objective brings representations of different views of the same concept closer and spreads representations of views from different concepts apart.

\subsubsection{Concept mask proposal.} We use masks to annotate fine-grained concepts explicitly. Assume a mask generator is given, as diagramed at the bottom of \Cref{fig:overview}; we first pass a reference view $x_r$, defined as the circumscribed rectangle crop of the union of two views, into the mask generator to obtain a set of concept masks --- $\mathcal{M}_r=\{m_i\}_{i=1}^K$, where $K$ is the number of concepts. Since the reference view contains both the query view and the key view, their concept masks $\mathcal{M}_q$ and $\mathcal{M}_k$ are immediately obtained if we restore them in the reference view. Then, we derive concept representations in both views by masked average pooling (MAP) with resized concept masks. Specifically, we compute $q^c = h\left(\verb|MAP|\left(f_5(x_q), m_c\right)\right)$ and $k^c$ similarly, where $\verb|MAP|\left(z, m\right)=\sum_{ij}m_{ij}\cdot z_{ij}/\sum_{ij}m_{ij}$, and $z \in \mathbb{R}^{CHW}$ denotes feature maps, $m \in \{0, 1\}^{HW}$ is a binary indicator for each concept. Here, only the shared concepts in both views are considered, \ie, $m_c \in \mathcal{M}_q \cap \mathcal{M}_k$. 

\begin{comment}
\subsubsection{Multi-level concepts.} Convolutional neural network would abstract different degrees of semantics across levels (layers). To enable ConCL to capture more diverse forms of concepts and understand both high-level and low-level semantics, we optimize a multi-level concept loss, formulated as:
\begin{equation}
    L_c^{all} = \sum_{i}^{l} \gamma_i L_c^{i},
\end{equation}
where $\gamma_i$ denotes the loss weight for the $i$-th level, and $l$ is the number of levels. For simplicity, we set $\gamma_i=2^{i-l}$. This gives us a set of decreasing weights from high-level to low-level.
\end{comment}

Our analysis hereafter focuses on 1) What makes the success of dense prediction pre-training? 2) What kind of concepts are good \textit{for pathology images}? Different answers to these two questions reveal the characteristics of pathology images and the disparity between natural and pathology images, as we explore in \Cref{sec:roadmap}. Below, we first introduce the benchmark pipeline and setups.

\subsection{Benchmark Pipeline}

Despite the extensive benchmarks in natural images for dense tasks, to our knowledge, such studies are unfortunately \textit{absent} in current works for pathology. Note that studying SSL methods in pathology images is still at an early stage. Most current works focus on employing image-level SSL methods for classification tasks. Orthogonal to theirs, we investigate a wider range of SSL methods for object detection and instance segmentation tasks, which are of high clinical value. We hope our work could provide useful data points and baselines for future work.

\subsubsection{Brief implementations and baseline settings.} Unless specified, our explorations use the following settings (more details in \cref{sec:add_implementations,sec:det_config}):
\begin{itemize}
    \item [\tiny$\bullet$] \textit{Pre-training codebase.} We use \verb|OpenSelfSup|\footnote{\url{https://github.com/open-mmlab/OpenSelfSup}} as our codebase since it incorporates various state-of-the-art self-supervised methods. All experiments are conducted within this codebase for integrity and fairness.
    \item [\tiny$\bullet$] \textit{Architecture.} We use ResNet-18 \cite{he2016deep} as the default backbone. For methods with MLP projection heads, \ie, MoCo-v2 \cite{chen2020improved}, SimCLR \cite{chen2020simple}, BYOL \cite{grill2020bootstrap}, PCL-v2 \cite{li2020prototypical}, and DenseCL \cite{wang2021dense}, we use the their default structures, but modify the numbers of input channels and hidden channels to 512 in accordance to ResNet-18. The number of output channels remains the same as the default.
    \item [\tiny$\bullet$] \textit{Hyper-parameter.} We change the queue length to 16384 in MoCo-v1/v2 \cite{he2020momentum,chen2020improved} and set the batch sizes of SimCLR \cite{chen2020simple} and BYOL \cite{grill2020bootstrap} to 1024. Other hyper-parameters remain the same as provided by the codebase, including data augmentation parameters and optimizer settings. Such settings should match their original proposals.
     \item [\tiny$\bullet$] \textit{Pre-training dataset.} We use NCT-CRC-HE-100K\cite{kather_jakob_nikolas_2018_1214456} dataset, referred to as NCT, for pre-training. It contains 100,000 non-overlapping patches extracted from hematoxylin and eosin (H\&E) stained colorectal cancer and normal tissues. All images are of size $224 \times 224$ at 0.5 MPP ($20\times$ magnification). We randomly choose 80\% of NCT to be the pre-training dataset.
     \item [\tiny$\bullet$] \textit{Transferring settings.} We use \verb|Detectron2| \cite{wu2019detectron2} as the detection codebase and the default hyperparameters. Unless otherwise specified, we use Mask-R-CNN \cite{he2017mask} detector with a feature pyramid network (FPN) head \cite{lin2017feature} as our base detector. For convenience and per common terminology \cite{he2019rethinking}, we define different $1\times$ fine-tuning schedules for two transferring datasets. For evaluation, We report the COCO-style metrics, \ie, mAP family.
    \item [\tiny$\bullet$] \textit{Transferring dataset.} We use two public datasets, the Gland segmentation in pathology images challenge (GlaS) dataset \cite{sirinukunwattana2017gland} and the Colorectal adenocarcinoma gland (CRAG) dataset \cite{graham2019mild}, and follow their official train/test splits for evaluation. GlaS \cite{sirinukunwattana2017gland} collects images of 775$\times$522 from H\&E stained slides with object-instance-level annotation; the images include both malignant and benign glands. CRAG \cite{graham2019mild} collects 213 H\&E stained images taken from 38 WSIs with a pixel resolution of 0.55$\mu$m/pixel at 20$\times$ magnification. Images are mostly of size 1512$\times$1516 with object-instance-level annotation. We study the performance of object detection and instance segmentation tasks.
\end{itemize}

\subsubsection{Experimental setup.} We pre-train all the methods on the NCT training set for 200 epochs. For ConCL pre-training, we warm up the model by optimizing instance contrastive loss (\cref{eq:moco_loss}) for the first 20 epochs and switch to concept contrastive loss (\cref{eq:concept_loss}). Then, we use the pre-trained backbones to initialize the detectors, fine-tune them on the training sets of transferring datasets, and test them in the corresponding test sets. %No validation set is used for both pre-training and transferring. 
Unless otherwise specified, we run all the transferring experiments 5 times and report the averaged performance.

%%%%%%%%%%%%%%%%%%%%%%%%%%%%%%%%%%%%%%%%%%%%%%%%%%%%%%%%%%%%%%%%%%%%%%%%%%%%%%%%%%%%%%%%%%%%%%%%%%%
\section{Towards Better Concepts: a Roadmap}
\label{sec:roadmap}

In this section, we first benchmark some popular state-of-the-art SSL methods for dense pathology tasks. Then, we start with DenseCL \cite{wang2021dense} and derive better concepts along the way, directed by the questions raised in the previous section.

\subsection{Benchmarking SSL methods for Dense Pathology Tasks}\label{sec:benchmark}

%################################################################################################## 
\begin{table}[t]\centering\resizebox{\textwidth}{!}{%
	\begin{tabular}{c|l|cc|cc|cc|cc}
& & 
  \multicolumn{4}{c|}{GlaS} &
  \multicolumn{4}{c}{CRAG} \\ & &
  \multicolumn{2}{c|}{Detect} &
  \multicolumn{2}{c|}{Segment} &
  \multicolumn{2}{c|}{Detect} &
  \multicolumn{2}{c}{Segment} \\
    \multirow{-2}{*}{Category} & \multirow{-2}{*}{Methods} & AP$^{bb}$   & AP$^{bb}_{75}$ & AP$^{mk}$   & AP$^{mk}_{75}$ & AP$^{bb}$  & AP$^{bb}_{75}$ & AP$^{mk}$  & AP$^{mk}_{75}$ \\ \shline
  \multirow{2}{*}{Baselines} 
    & {\color{darkgray}Rand. Init.} 
            & 49.8 & 57.3 & 52.1 & 60.7 & 51.1 & 57.0 & 50.6 & 57.3 \\
	& Supervised & 50.2 & 56.9 & 53.2 & 62.1 & 49.2 & 55.2 & 49.4 & 55.0 \\ \hline
\multirow{6}{*}{\begin{tabular}[x]{@{}c@{}}\cref{sec:benchmark}\\Prior SSL arts\end{tabular}}
    & SimCLR\cite{chen2020simple} & 50.7 & 56.9  & 53.6 & 62.7  & 49.2  & 54.8  & 49.1 & 54.7 \\
	& BYOL\cite{grill2020bootstrap}  & 50.9 & 57.7  & 53.9  & 62.6  & 49.9  & 55.8  & 49.3  & 55.3 \\
	& PCL-v2$^\dagger$ \cite{li2020prototypical} & 49.4  & 55.9  & 51.9  & 61.0  & 51.0  & 56.6  & 50.5  & 56.7 \\
	& MoCo-v1\cite{he2020momentum} & 50.0  & 56.2  & 52.1 & 59.9  & 47.2 & 51.1  & 47.5 & 52.0 \\
	& MoCo-v2\cite{chen2020improved} & 52.3  & 60.0  & 55.3  & 65.0  & 50.0  & 55.7  & 50.3  & 56.8 \\
    & DenseCL\cite{wang2021dense} & 53.9 & 62.0  & 56.5  & 66.2  & 52.3  & 58.2  & 52.2  & 59.8 \\ \hline
\multicolumn{10}{l}{\textit{Our differently instantiated ConCLs:}}\\
\multirow{3}{*}{\begin{tabular}[x]{@{}c@{}}\cref{sec:grid_concept}\\Grid concepts\end{tabular}}
	& (1) g-ConCL(s=3) & 54.9 & 64.1 & 57.1 & 66.3 & 55.4 & 62.3 & 54.4 & 62.0 \\
    & (2) g-ConCL(s=5) & 55.4 & 65.2 & 57.4 & 67.2 & 55.5 & 62.7 & 54.6 & 62.2 \\
    & (3) g-ConCL(s=7) & 54.9 & 63.8 & 57.0 & 66.5 & 55.3 & 62.5 & 54.7 & 62.6 \\ \hline
\multirow{3}{*}{\begin{tabular}[x]{@{}c@{}}\cref{sec:natural_prior}\\Natural-image\\priors concepts\end{tabular}}
    & (4) fh-ConCL(s=50) & 55.8 & 65.6 & 58.3 & 68.8 & 54.8 & 60.7 & 54.1 & 60.7 \\
    & (5) fh-ConCL(s=500) & 56.2 & 65.9 & 57.7 & 67.9 & 54.7 & 61.9 & 53.8 & 60.5 \\
    %& (6) fh-ConCL(s=1000) & 56.3 & 65.9 & 58.4 & 68.4 & 54.9 & 61.8 & 54.2 & 61.3 \\
    & (6) bas-ConCL & 56.1 & 66.1	& 58.1	& 68.1 & 54.2 & 61.1 &	53.4 & 60.8 \\ \hline
\multirow{3}{*}{\begin{tabular}[x]{@{}c@{}}\cref{sec:wsi_prior}\\Bootstrapped\\concepts\end{tabular}} & & & & & & \\
    & (7) b-ConCL($f_4$)   & \textbf{56.8} & \textbf{66.2} &\textbf{58.7} & \textbf{68.9} & 55.1 & 62.2 & 54.1 & 61.4	      \\ 
    & (8) b-ConCL($f_5$) & 56.1 & 65.6 & 57.8 & 67.7 & \textbf{56.5} & \textbf{63.3} & \textbf{55.3} & \textbf{62.9} 
\end{tabular}
}
%##################################################################################################

\caption{\textbf{Main results of object detection and instance segmentation.} All models are pre-trained for 200 epochs on the NCT dataset \cite{kather_jakob_nikolas_2018_1214456} and fine-tuned on GlaS \cite{sirinukunwattana2017gland} and CRAG \cite{graham2019mild} with ResNet-18 Mask-RCNN-FPN under a $1\times$ schedules. The results are averaged over 5 independent trials. $^\dagger$ PCL-v2 is trained using the officially released code. AP$^{bb}$: bounding box mAP, AP$^{mk}$: mask mAP.}
\label{tab:main_results}
\end{table}
%##################################################################################################
\subsubsection{Benchmark results.} \Cref{tab:main_results} (baselines and prior SSL arts) shows the transferring performance for GlaS dataset (left columns) and CRAG dataset (right columns), respectively. We report results using 200-epoch pre-trained models and a $1\times$ fine-tuning schedule. On the GlaS dataset \cite{sirinukunwattana2017gland}, we observe that the gap between training from randomly initialized models and training from supervised pre-trained models is relatively smaller compared to those in the natural image domain \cite{chen2021exploring,chen2020improved,grill2020bootstrap,chen2020simple}. Nonetheless, state-of-the-art SSL methods all exceed supervised pre-training, meeting the same expectation as in natural images. Yet, on the CRAG dataset \cite{graham2019mild}, most of the pre-trained models, including both the self-supervised ones and the supervised one, fail to achieve competitive performance compared to training from randomly initialized weights. The only exception is DenseCL \cite{wang2021dense}, a dense contrasting method. 
 	
Overall, among the image-level SSL methods, MoCo-v2 \cite{chen2020improved} performs the best in GlaS and the second-best in CRAG. Built upon MoCo-v2 and enhanced by dense contrasting, DenseCL \cite{wang2021dense} achieves the best results in both datasets. It should be emphasized that DenseCL \cite{wang2021dense} gets + 1.6 AP$^{bb}$ for GlaS by using grid-level contrasting. This demonstrates the importance of designing dense pre-training frameworks when transferring to dense tasks since all the stragglers are only optimized for image-level representations.

Thus, we here conclude \textit{dense contrasting matters}.

%##################################################################################################
\begin{figure}[t]\centering
\includegraphics[width=\textwidth]{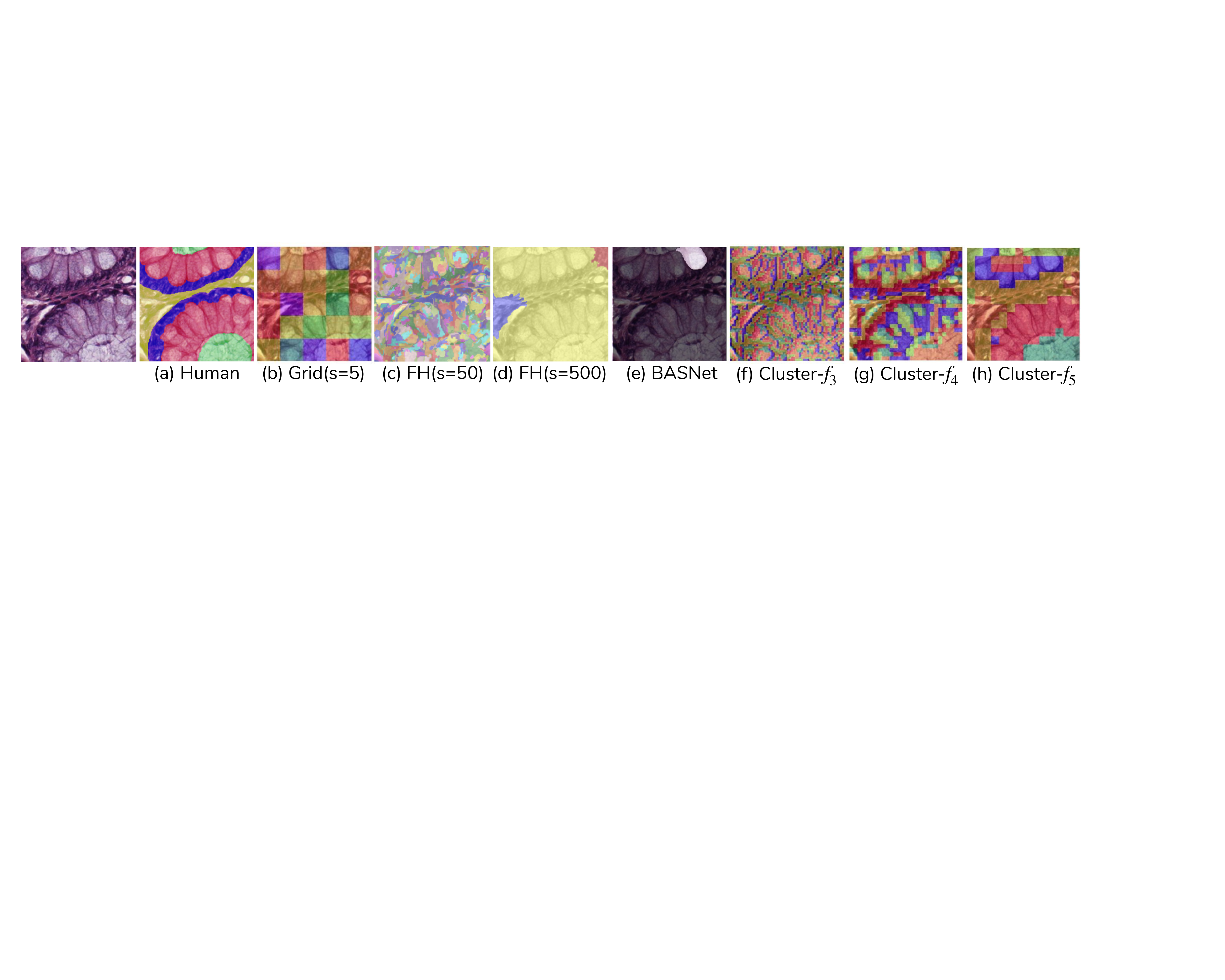} 
\caption{\textbf{Concept descriptors.} (a) Tissue concept illustration. (b) Grid concepts (s: grid number). (c-d) FH concepts (s: scale). (e) Binary saliency concepts, obtained from BASNet \cite{qin2019basnet}. (f-h) Clustering concepts ($f_i$: ResNet output stage). The image is resized to $448 \times 448$ for better visualization.}
\label{fig:cluster}
\end{figure}
%##################################################################################################

\subsection{Correspondence matters}\label{sec:grid_concept}

From the previous section, we find dense contrasting is favored in both natural and pathology images, where DenseCL \cite{wang2021dense} all achieves top performance. The next question is: \textit{can we improve the dense contrasting framework?} To answer it, we first summarize the overall pipeline of DenseCL \cite{wang2021dense}. DenseCL computes the dense representations of two views without global average pooling, \ie, $f_5(x_q), f_5(x_k)$, and passes them to a dense projection head to obtain final grid features of size $\mathbb{R}^{128\times7\times7}$. Then it sets the most similar (measured by cosine similarity) grids in two views as positive pairs. As such, the correspondence of positive pairs is learned. However, the reliability of learned correspondence remains questionable and would affect the quality of learned representations.

To address that, we instantiate DenseCL \cite{wang2021dense} in ConCL by regarding the grid-prior as a form of concept, as shown in \Cref{fig:cluster}-(b). We denote this ConCL instance as g-ConCL. Compared with DenseCL \cite{wang2021dense} (learned matching), ConCL naturally restores the positive correspondence from a reference view (precise matching, \cref{fig:overview}-$x_r$) which is more reliable. \Cref{tab:main_results}-(1-3) compares the original DenseCL \cite{wang2021dense} and ConCL-instantiated g-ConCL. The results indicate that g-ConCL with precise correspondence can boost DenseCL \cite{wang2021dense} by a large margin. Even with the simplest form of concepts, g-ConCL already has topped entries above it in \Cref{tab:main_results}. We believe other dense pre-training methods that learn the matching between grids, \eg, \textit{Self}-EMD \cite{liu2020self}, should perform similarly to DenseCL \cite{wang2021dense}, and g-ConCL could outperform them.

Thus, we here conclude that \textit{correspondence matters}.

\subsection{Natural Image Priors in Pathology Images}\label{sec:natural_prior}

ConCL is a general framework for using masks as supervision to discriminate concepts. Some previous works in natural image \cite{zhao2021contrastive,henaff2021efficient,zhao2020distilling,van2021unsupervised,wang2021exploring} also combines masks with contrastive learning, where the masks are provided by ground truth annotation \cite{zhao2021contrastive,wang2021exploring,henaff2021efficient}, or supervised/unsupervised pseudo-mask generation \cite{henaff2021efficient,zhao2020distilling,van2021unsupervised}. The mask generators can be graph-based (\eg, Felzenszwalb-Huttenlocher algorithm \cite{felzenszwalb2004efficient}), MCG \cite{arbelaez2014multiscale}, or other saliency detection models \cite{qin2019basnet,nguyen2019deepusps} trained on designated natural image datasets. However, those methods werer originally proposed for nature images, and their success for pathology images remains unknown. 

Here we instantiate ConCL by using Felzenszwalb-Huttenlocher (FH) algorithm \cite{felzenszwalb2004efficient} and BASNet \cite{qin2019basnet} as concept generators, dubbed as fh-ConCL and bas-ConCL, respectively. FH \cite{felzenszwalb2004efficient} is a conventional graph-based segmentation algorithm that relies on local neighborhoods, while BASNet \cite{qin2019basnet} is a deep neural network pre-trained on a curated saliency detection dataset, which only contains daily natural objects. We use these two as representatives to study if these natural image priors win twice in both natural and pathology images. 

For implementations, we use the FH algorithm in the scikit-image package and set both ``scale'' and ``size'' hyper-parameters to $s$. We use the pre-trained BASNet provided by \cite{qin2019basnet}. \Cref{fig:cluster}-(c-e) shows some examples. \Cref{tab:main_results} reports the results.

It is not surprising that the BASNet \cite{qin2019basnet} cannot generate decent concept masks (\cref{fig:cluster}-(e)) for pathology images since it is pre-trained on  curated saliency detection datasets. What is surprising is that bas-ConCL does yield satisfactory results (\cref{tab:main_results}-(6)). Similar observations are also found in fh-ConCLs (\cref{tab:main_results}-(4,5)) that though the generated concept masks are coarse-grained, the resulted transferring performances are unexpectedly good. After inspecting more examples, we find that the generated masks maintain high coherence and integrity despite their coarse-grained nature. That said, each concept contains semantic-consistent objects or textures. For example, \Cref{fig:cluster}-(d,e) can be seen as special cases of \Cref{fig:cluster}-(a) that merge fine-grained semantics with coarse-grained ones. This property makes the major difference between fh-/bas-ConCLs and g-ConCLs, where the grid-concepts are less likely to have coherent semantics.

Thus, we here conclude that \textit{coherence matters} and natural image priors also work in pathology images, though they mostly provide coarse-grained concepts.

\subsection{Pathology Image Priors in Pathology Images}\label{sec:wsi_prior}

Can we obtain concept masks away with natural image priors? External dependency is not always wanted and sometimes may fail to provide the desired masks (\eg, \cref{fig:cluster}-(e)). We thus task ourselves to find a dependency-free concept proposal method. One of the key characteristics inherent in pathology images is that they have rich low-level patterns and tissue structures. Can we use that prior instead?

\Cref{fig:cluster}-(f-h) shows the clustering visualization from intermediate feature maps generated by a 10-epoch warmed-up MoCo-v2 \cite{chen2020improved}. Thanks to the rich structural patterns in pathology images, we find that simply clustering over the feature maps provided by a barely trained model can already generate meaningful structural concept proposals. We thus build upon this ``free lunch'' and use a ``bootstrap your own \textit{perception}'' mechanism that is similar to the ``bootstrap your own latent'' mechanism in BYOL \cite{grill2020bootstrap}. ConCL generates concept proposals from the momentum key encoder's perception while simultaneously improving and refining it via the online query encoder, leading to a ``bootstrapping'' behavior. Thus, we denote such ConCL as bootstrapped-ConCL (b-ConCL). We provide an additional introduction to BYOL and ``bootstrapping'' in \Cref{sec:bootstrap}.

\subsubsection{b-ConCL.} The concept generator is now instantiated as a KMeans grouper. We first pass the reference view $x_r$ to the key encoder to obtain a reference feature map from ResNet stage-$i$: $f_{i}(x_r) \in \mathbb{R}^{CHW}$. Then, we apply K-Means to group $K$ underlying concepts. b-ConCL relies on neither external segmentation algorithms nor designated saliency detection models for natural images.

Our default setting is $K=8$, and clustering from $f_4$ or $f_5$. We postpone the study of hyper-parameters, \ie, the number of clusters in KMeans, and the clustering stage $f_i$ to \Cref{sec:ablation} and report the main results in \Cref{tab:main_results}-(7,8). We find b-ConCL tops other entries. Compared to MoCo-v2 \cite{chen2020improved}, our direct baseline, b-ConCL outperforms it by +4.5 AP$^{bb}$ and +3.1 AP$^{mk}$. Moreover, b-ConCL obtains more gains in terms of AP$_{75}$ (+6.2 AP$^{bb}_{75}$, +3.7 AP$^{mk}_{75}$) compared to MoCo-v2 \cite{chen2020improved}, which means it improves MoCo-v2 \cite{chen2020improved} by more accurate bounding box regression and instance mask prediction. This aligns with our motivation for ConCL since discriminating local concepts helps shape object borders. 

\subsubsection{Closing remarks.} So far, we have included: i) dense contrasting matters; ii) correspondence matters; iii) coherence matters; iv) natural image priors, though they might only provide coarse-grained concepts, work in pathology images as well; and find v) a randomly initialized or barely trained convolutional neural network, thanks to the rich low-level patterns in pathology images and good network initialization, can generate good proposals that are \textit{dense}, \textit{fine-grained} and \textit{coherent}, as shown in \Cref{fig:cluster}. Though the coarse-grained concepts generated from natural image priors could also help tasks in our studied benchmarks, they might underperform when a fine-grained dense prediction task is given, which we leave for future work.

We hope our closing remarks could be intriguing and guide future works in designing dense pre-training methods for pathology images and beyond. 

%%%%%%%%%%%%%%%%%%%%%%%%%%%%%%%%%%%%%%%%%%%%%%%%%%%%%%%%%%%%%%%%%%%%%%%%%%%%%%%%%%%%%%%%%%%%%%%%%%%
\section{More Experiments}

In the previous section, we have explored how we can obtain concepts, what concepts are good, and find b-ConCL to be the best. We here conduct more experiments to study b-ConCL further. Some qualitative visual comparisons are in \Cref{sec:more_results}.

\subsection{Robustness to Transferring Settings}
 
 %##################################################################################################
\begin{table}[t]
\centering
\setlength{\tabcolsep}{1mm}{
\resizebox{.75\textwidth}{!}{
\begin{tabular}{c|c|ll|ll}
\multirow{2}{*}{ Detector } & \multirow{2}{*}{Pretrain}	&\multicolumn{2}{c|}{GlaS Detection}	&\multicolumn{2}{c}{CRAG Detection} \\
\cline{3-6} 
&&{\centering AP$^{bb}$}	&	AP$^{bb}_{75}$	&AP$^{bb}$	&AP$^{bb}_{75}$\\
\shline
\multirow{4}{*}{MaskRCNN+C4}&
Rand. Init.	&52.9	&59.9	&49.4	&54.2 \\
&Supervised
	&49.1{\footnotesize \color{darkgray}{(-3.8)}}
	&55.1{\footnotesize \color{darkgray}{(-4.8)}}
	&46.1{\footnotesize \color{darkgray}{(-3.3)}}
	&50.6{\footnotesize \color{darkgray}{(-2.3)}} \\
&MoCo-v2 \cite{chen2020improved}
	&53.6{\footnotesize \color{darkpastelgreen} (+0.7)}
	&61.8{\footnotesize \color{darkpastelgreen} (+1.9)}
	&48.3{\footnotesize \color{darkgray} (-1.1)}
	&52.6{\footnotesize \color{darkgray} (-1.6) }\\
&\textbf{b-ConCL}
	&55.8{\footnotesize \color{darkpastelgreen} (+2.9)}
	&63.6{\footnotesize \color{darkpastelgreen} (+3.7)}
	&49.8{\footnotesize \color{darkpastelgreen} (+0.4)}
	&54.3{\footnotesize \color{darkpastelgreen} (+0.1)} \\
\hline
\multirow{4}{*}{MaskRCNN+FPN}
&Rand. Init.	&49.8	&57.3	&51.1	&57.0 \\
&Supervised
	&50.2{\footnotesize \color{darkpastelgreen} (+0.4)}
	&56.9{\footnotesize \color{darkgray}{(-0.4)}}
	&49.2{\footnotesize \color{darkgray}{(-1.9)}}
	&55.2{\footnotesize \color{darkgray}{(-1.8)}} \\
&MoCo-v2 \cite{chen2020improved}
	&52.3{\footnotesize \color{darkpastelgreen} (+2.5)}
	&60.0{\footnotesize \color{darkpastelgreen} (+2.7)}
	&50.0{\footnotesize \color{darkgray}{(-1.1)}}
	&55.7{\footnotesize \color{darkgray}{(-1.3)}} \\
&\textbf{b-ConCL}
	&56.8{\footnotesize \color{darkpastelgreen}(+7.0)}
	&66.2{\footnotesize \color{darkpastelgreen}(+8.9)}
	&55.1{\footnotesize \color{darkpastelgreen} (+4.0)}
	&62.2{\footnotesize \color{darkpastelgreen} (+5.2)} \\
\hline
\multirow{4}{*}{RetinaNet}
&Rand. Init.	&46.4	&51.0	&45.2	&47.6 \\
&Supervised
	&44.7{\footnotesize \color{darkgray}{(-1.7)}}
	&48.4{\footnotesize \color{darkgray}{(-2.6)}}
	&43.1{\footnotesize \color{darkgray}{(-2.1)}}
	&44.8{\footnotesize \color{darkgray}{(-2.8)}} \\
&MoCo-v2 \cite{chen2020improved}
	&47.2{\footnotesize \color{darkpastelgreen}(+0.8)}
	&50.9{\footnotesize \color{darkgray}{(-0.1)}}
	&43.1{\footnotesize \color{darkgray}{(-2.1)}}
	&43.8{\footnotesize \color{darkgray}{(-3.8)}} \\
&\textbf{b-ConCL}
	&52.6{\footnotesize \color{darkpastelgreen} (+6.2)}
	&58.6{\footnotesize \color{darkpastelgreen} (+7.6)}
	&48.4{\footnotesize \color{darkpastelgreen}(+3.2)}
	&51.9{\footnotesize \color{darkpastelgreen} (+4.3)} \\
\hline
\end{tabular}}
}
\caption{\textbf{Detection performance using different detectors.} All methods are pre-trained for 200 epochs and fine-tuned with $1\times$ schedule. Results are averaged over 5 trials.}
\label{tab:diff_detectors}
\end{table}
%##################################################################################################

\subsubsection{Transferring with different detectors.} Here we investigate the transferring performance with other detectors, \ie, Mask-RCNN-C4 (C4) \cite{ren2015faster} and RetinaNet \cite{lin2017focal}. RetinaNet is a single-stage detector. It uses ResNet-FPN backbone features as Mask-RCNN-FPN but directly generates predictions without region proposal \cite{ren2015faster}. C4 detector adopts a similar two-stage fashion as Mask-RCNN but uses the outputs of the 4-th residual block as backbone features and re-targets the 5-th block to be the detection head instead of building a new one. These three representative detectors evaluate pre-trained models under different detector architectures. All detectors are fine-tuned from 200-epoch pre-trained models with $1\times$ schedule for 5 independent trials. Results together with Mask-RCNN-FPN's are shown in \Cref{tab:diff_detectors}. b-ConCL performs the best with all three detectors in both datasets. Notably, training from scratch (Rand. Init.) is one of the top competitors when the C4 detector is used. We conjecture that the pre-trained models are possibly overfitted to their pretext tasks in their 5-th blocks and thus are harder to be tuned than a randomly initialized 5-th block. In CRAG detection, only b-ConCL pre-trained models consistently outperform randomly initialized models. In addition, the most significant gap between MoCo-v2\cite{chen2020improved} and b-ConCL is found in the RetinaNet detector \cite{lin2017focal}. As also noted by \cite{liu2020self}, RetinaNet \cite{lin2017focal} is a single-stage detector, where the local representations from the backbone become more important than other two-stage detectors since results are directly predicted from them. b-ConCL is tasked to discriminate local concepts, and subsequently, the learned representations could be better than other pre-training methods here. 

\subsubsection{Transferring with different schedules.} Fine-tuning with longer schedules may improve downstream task performance. To investigate if b-ConCL's lead could persist with longer fine-tuning, we fine-tune Mask-RCNN-FPN with $0.5\times$, $1\times$, $2\times$, $3\times$, and $5\times$ schedules. \Cref{tab:diff_lrs} shows the results. b-ConCL maintains its noticeable gains in longer schedules in both datasets, \eg, b-ConCL achieves 56.2 mAP with a $0.5\times$ schedule, which is better than MoCo-v2 \cite{chen2020improved} with a $5\times$ schedule but costs 10 $\times$ less fine-tuning time. Similar observations are also found in CRAG, where the gap between b-ConCL and MoCo-v2 \cite{chen2020improved} becomes larger (see $\mathrm{\Delta}$ row). Together, these results confirm b-ConCL's superiority across different fine-tuning schedules.

%##################################################################################################
\begin{table}[b]
\centering
\resizebox{0.7\linewidth}{!}{%
\begin{tabular}{c|ccccc|ccccc}
\multirow{3}{*}{Method} & \multicolumn{5}{c|}{GlaS dataset} & \multicolumn{5}{c}{CRAG dataset} \\ \cline{2-11} 
						& \multicolumn{5}{c|}{Fine-tuning schedule} & \multicolumn{5}{c}{Fine-tuning schedule}  \\ \cline{2-11}
                        & $0.5\times$   & $1\times$     & $2\times$     & $3\times$     & $5\times$ 
                        & $0.5\times$   & $1\times$     & $2\times$     & $3\times$     & $5\times$    \\ \shline
Rand. Init.              & 49.1          & 49.8          & 51.4          & 51.8          & 52.7     
						& 50.2          & 51.1          & 51.9          & 52.4          & 52.8     \\
Supervised              & 48.6          & 50.2          & 51.4          & 52.7          & 54.0     
						& 50.0          & 49.2          & 50.5          & 50.1          & 50.3     \\
MoCo-v2\cite{chen2020improved}  & 51.4          & 52.3          & 53.7          & 54.2          & 55.7 
						& 50.2          & 50.0          & 50.2          & 50.8          & 51.8         \\ \hline
\textbf{b-ConCL}          & \textbf{56.2} & \textbf{56.8} & \textbf{57.7} & \textbf{58.3} & \textbf{59.0} 
						& \textbf{54.8} & \textbf{55.1} & \textbf{55.4} & \textbf{55.6} & \textbf{56.0} \\
$\mathrm{\Delta}$		& {\small +4.8} & {\small +4.5} & {\small +4.0} & {\small +4.1} & {\small +3.3}
						& {\small +4.6} & {\small +5.1} & {\small +5.2} & {\small +4.8} & {\small +4.2}

\end{tabular}%
}
\caption{\textbf{Detection performance under different fine-tuning schedules.} Results other than $1\times$ schedule are averaged over 3 runs. $\mathrm{\Delta}$ row shows b-ConCL's improvement over MoCo-v2. We report AP$^{bb}$ here.}
\label{tab:diff_lrs}
\end{table}
 %##################################################################################################

\subsection{Ablation Study}\label{sec:ablation}

In this section, we ablate the key factors in b-ConCL. Our default setting clusters $K=8$ concepts from ResNet stage-4 ($f_4(\cdot)$). Since b-ConCL is built on MoCo-v2 \cite{chen2020improved}, we use it as our direct baseline for comparisons.

\subsubsection{Concept loss weight $\lambda$.}\label{sec:ablation_loss_weight} We here study the generalized concept contrastive loss: $L = (1-\lambda) L_q + \lambda L_c$, where $\lambda \in [0, 1]$ is a concept loss weight parameter. It shows a natural way to combine concept contrastive loss with instance contrastive loss. We start by asking whether instance contrastive loss is indispensable during the training process of b-ConCL. We alter the concept loss weight $\lambda$, and \Cref{tab:loss_weight} reports the results. We see a monotonically increasing performance as $\lambda$ increases in both datasets, which emphasizes the importance of concept loss. When no warm-up is used (last row in \cref{tab:loss_weight}), only a slight performance drop is observed, meaning that warm-up is not the key component of b-ConCL. Warming-up with instance loss (\cref{eq:concept_loss}) is a special case of b-ConCL, where at the early training stage, each instance is regarded as a concept, and we then gradually increase the number of concepts as training goes on. Thus, the overall findings in this ablation support b-ConCL's advance over MoCo-v2 \cite{chen2020improved}.
   
%##################################################################################################
% overall table of all ablations
\begin{table*}[t]
\centering
%#################################################
% Concept loss weight
%#################################################
\subfloat[
\textbf{Concept loss weight}.\label{tab:loss_weight}]{
\centering
\begin{minipage}{0.4\linewidth}{\begin{center}
\tablestyle{4pt}{1.05}
\resizebox{.9\linewidth}{!}{%
\begin{tabular}{x{18}x{24}x{24}|x{24}x{24}}
\multirow{2}{*}{$\lambda$} & \multicolumn{2}{c|}{GlaS} & \multicolumn{2}{c}{CRAG} \\
        & AP$^{bb}$ & AP$^{bb}_{75}$ & AP$^{bb}$ & AP$^{bb}_{75}$  \\
\shline
\gc{0.0}& \gc{52.3} & \gc{60.0} & \gc{50.0} & \gc{55.7}		 \\ 
0.1     & 53.6      & 61.1      & 50.5      & 55.9       \\
0.3     & 53.6      & 61.8      & 51.7      & 57.1       \\
0.5     & 53.6      & 61.8      & 51.3      & 57.0       \\
0.7     & 55.2      & 64.1      & 53.1      & 59.9       \\
0.9     & 56.0      & 65.1      & 53.6      & 59.6    \\
\baseline{1.0}   &\textbf{56.8}&\textbf{66.2}&\textbf{55.1}&\textbf{62.2} \\\hline
1.0\textbackslash w. & 56.1		 &\textbf{66.2}		   & 54.0	     & 60.6 \\
\end{tabular}}
\end{center}}\end{minipage}}
%#################################################
% Number of concepts
%#################################################
\subfloat[\textbf{Number of concepts}.\label{tab:num_concepts}]{
\begin{minipage}{0.4\linewidth}{\begin{center}
\tablestyle{4pt}{1.05}
\resizebox{.9\linewidth}{!}{%
\begin{tabular}{x{18}x{24}x{24}|x{24}x{24}}
\multirow{2}{*}{$K$}& \multicolumn{2}{c|}{GlaS} & \multicolumn{2}{c}{CRAG} \\
        & AP$^{bb}$ & AP$^{bb}_{75}$ & AP$^{bb}$ & AP$^{bb}_{75}$  \\
\shline
\gc{1}& \gc{52.3} & \gc{60.0} & \gc{50.0} & \gc{55.7}		 \\ 
2      & 54.5      & 64.1      & 52.9      & 60.1       \\
4      & 55.6      & 64.7      & 53.4      & 59.7       \\
6      & 56.3      & 65.1      & 53.7      & 60.2       \\
\baseline{8}& 56.8      & \textbf{66.2}      & \textbf{55.1}      & \textbf{62.2}       \\
10      & 57.0      & 66.0      & \textbf{55.1}      & 61.0       \\
12      & \textbf{57.4}      & \textbf{66.2}      & 54.2      & 60.1      \\
16      & 55.7      & 65.3      & 54.5      & 61.3 
\end{tabular}}
\end{center}}\end{minipage}}
\\
\centering
%#################################################
% Clustering stage
%#################################################
\subfloat[
\textbf{Clustering stage.}\label{tab:cluster_stage}]{

\begin{minipage}{0.4\linewidth}{\begin{center}
\tablestyle{4pt}{1.05}
\resizebox{.9\linewidth}{!}{%
\begin{tabular}{x{15}x{24}x{24}|x{24}x{24}}
\multirow{2}{*}{$K$}& \multicolumn{2}{c|}{GlaS} & \multicolumn{2}{c}{CRAG} \\
        & AP$^{bb}$ & AP$^{bb}_{75}$ & AP$^{bb}$ & AP$^{bb}_{75}$  \\
\shline
\gc{None}          & \gc{52.3} & \gc{60.0} & \gc{50.0} & \gc{55.7}		 \\ 
$f_1(\cdot)$       & 55.0             & 65.1          & 53.3          & 60.0              \\
$f_2(\cdot)$       & 55.0             & 64.7          & 53.7          & 60.4          \\
$f_3(\cdot)$       & \underline{56.2} & \textbf{66.4} & 53.0          & 59.6          \\
\baseline{$f_4(\cdot)$} & \textbf{56.8}    & \underline{66.2} & \underline{55.1} & \underline{62.2} \\
$f_5(\cdot)$       & 56.1             & 65.6             & \textbf{56.5} & \textbf{63.3} \\
\end{tabular}}
\end{center}}\end{minipage}}
%#################################################
% Backbone capacities
%#################################################
\subfloat[
\textbf{Backbone capacities.}\label{tab:backbone_caps}]{
\begin{minipage}{0.4\linewidth}{\begin{center}
\tablestyle{4pt}{1.05}
\resizebox{\linewidth}{!}{%
\begin{tabular}{y{40}x{24}x{24}|x{24}x{24}}
\multicolumn{5}{c}{GlaS Detection}\\\hline
\multirow{2}{*}{Pretrain}& \multicolumn{2}{c|}{\baseline{ResNet-18}} & \multicolumn{2}{c}{ResNet-50} \\
        & AP$^{bb}$ & AP$^{bb}_{75}$ & AP$^{bb}$ & AP$^{bb}_{75}$  \\
\shline
Rand.	&49.8	&57.3	&49.9	&56.1 \\
Sup.  &50.2	&56.9	&47.9	&54.2  \\
\gc{MoCo.v2}     &52.3	&60.0	&53.1	&60.5 \\ \hline
\baseline{b-ConCL}	&\textbf{56.8}	&\textbf{66.2}	& \textbf{57.0}	&\textbf{65.9}\\
\multicolumn{5}{l}{~}
\end{tabular}}
\end{center}}\end{minipage}}
%#################################################
\caption{\textbf{Ablation Study.} We study the effect of different hyper-parameters to b-ConCL. Default settings are marked in \colorbox{baselinecolor}{gray} and MoCo-v2 baselines are marked by \gc{gray}. In (a), ``\textbackslash w.'' means no warm-up.}
\label{tab:ablations} 
\end{table*}
%##################################################################################################

\subsubsection{Number of concepts $K$.} Here, we investigate how the number of concepts clustered during pre-training affects performance in downstream tasks. We report the results of different $K$ in \Cref{tab:num_concepts}. b-ConCL performs reasonably well when $K>=4$, with most of performance peaking at $K=8$. This demonstrates the robustness of b-ConCL to the choice of $K$. Note that the best performance for the GlaS dataset is higher than our default setting and outperforms all entries in \Cref{tab:main_results}, showing the potential room for b-ConCL. 

\subsubsection{Where to group $f_i(\cdot)$.} b-ConCL groups concepts from a model's intermediate feature maps. Our default setting uses feature maps from stage-4 of a ResNet \cite{he2016deep}, denoted as $f_4(\cdot)$. We now ablate this choice in \Cref{tab:cluster_stage}. Clustering concepts from $f_4(\cdot)$ and $f_5(\cdot)$ works similarly well across two datasets. We choose $f_4(\cdot)$ as the default since it achieves top two performance in both datasets under both metrics. Besides, b-ConCL exceeds MoCo-v2 \cite{chen2020improved}, whichever stage it groups concepts from. This again confirms the effectiveness and robustness of b-ConCL. See \Cref{fig:cluster} as a reference for the cluster results from parts of these stages.

\subsubsection{Longer pre-training.} We compare the pre-training efficiency of different SSL methods w.r.t. training epochs in \Cref{fig:longer_pretraining1}-(b,c) with the numerical results in \Cref{sec:more_results}. Interestingly, we find SimCLR \cite{chen2020simple} and BYOL \cite{grill2020bootstrap} fail to benefit from longer pre-training. This shows the disparity between pathology image data and natural image data. In the latter field, a monotonically increasing performance in downstream tasks is usually observed \cite{henaff2021efficient,he2020momentum,grill2020bootstrap,caron2020unsupervised,chen2020simple}. For MoCo-v1/v2 \cite{he2020momentum,chen2020improved}, DenseCL \cite{wang2021dense} and our b-ConCL, we observe the performance consistently improves as the pre-training epoch increases in GlaS dataset \cite{sirinukunwattana2017gland}. Note that, the 200-epoch pre-trained b-ConCL surpasses the 800-epoch pre-trained MoCo-v2 \cite{chen2020improved}, and DenseCL\cite{wang2021dense} by a large margin (\cref{fig:longer_pretraining1}-(b)). In the CRAG dataset, we observe all pre-training methods saturate and achieve the best performance in around 200-epoch and 400-epoch. Among them, b-ConCL is still the best (\cref{fig:longer_pretraining1}-(c)). Especially, the 100-epoch ConCL has already outperformed all the other competitors regardless pre-training epochs (\cref{fig:longer_pretraining1}-(b)).

\subsubsection{Larger model capacity.} \Cref{tab:backbone_caps} shows the results of using a larger backbone, ResNet-50. b-ConCL maintains its leading position. For consistency to the previous ablation, a $1\times$ schedule is also used here, which could put ResNet-50 at a disadvantage since it has more parameters to tune in a relatively short schedule.

%%%%%%%%%%%%%%%%%%%%%%%%%%%%%%%%%%%%%%%%%%%%%%%%%%%%%%%%%%%%%%%%%%%%%%%%%%%%%%%%%%%%%%%%%%%%%%%%%%%
\section{Conclusion and Broader Impact}

In this work, we have benchmarked some of the current SSL methods for dense tasks in pathology images and presented the ConCL framework. Along our exploration, we have distilled several key components to the success of transferring to dense tasks: i) dense contrasting matters, ii) correspondence matters, iii) coherence matters, and more. Finally, we ended up with a dependency-free concept generator that directly bootstraps the underlying concepts inherent in the data and learns from \textit{scratch}. It was shown to be robust and competitive.

While our initial results are presented only for pre-training and fine-tuning, many applications could embrace ConCL. One example is to combine it with few-shot detection or segmentation, where clustering from feature arrays can be an approach for mining latent objects. Another example can be semi-supervised learning, where ConCL can be used as an additional branch for unlabeled data. \textit{Beyond} pathology image analysis, we also hope ConCL would help in speech or tabular data, where little priors can be used. Unsupervised clustering in representation space is likely to be modality-agnostic. Learned by using contrastive learning and clustering, fine-grained ``concepts'' could also be mined from those data modalities.

\subsubsection*{Acknowledgments}
This work is done by Jiawei Yang during an internship at Tencent AI Lab. Hanbo Chen and Jianhua Yao are the corresponding authors of this paper.

%%%%%%%%%%%%%%%%%%%%%%%%%%%%%%%%%%%%%%%%%%%%%%%%%%%%%%%%%%%%%%%%%%%%%%%%%%%%%%%%%%%%%%%%%%%%%%%%%%%
\bibliographystyle{splncs04}
\bibliography{arxiv}
\clearpage

%%%%%%%%%%%%%%%%%%%%%%%%%%%%%%%%%%%%%%%%%%%%%%%%%%%%%%%%%%%%%%%%%%%%%%%%%%%%%%%%%%%%%%%%%%%%%%%%%%%
\appendix
%%%%%%%%%%%%%%%%%%%%%%%%%%%%%%%%%%%%%%%%%%%%%%%%%%%%%%%%%%%%%%%%%%%%%%%%%%%%%%%%%%%%%%%%%%%%%%%%%%%
\section{Bootstrapping}\label{sec:bootstrap}

\setcounter{figure}{0}
\setcounter{footnote}{0}
\renewcommand{\thefigure}{A.\arabic{figure}}

\subsubsection{Bootstrapping.} As stated in BYOL \cite{grill2020bootstrap}: ``the term \textit{bootstrap} is used in its idiomatic sense rather than the statistical sense,'' \eg, DeepCluster \cite{caron2018deep} uses bootstrapping on previous versions of its representation to produce targets (cluster indices) for the next representation. Methods based on self-training or pseudo-labels are also considered bootstrapping since they use information from previous steps to provide \emph{targets} for next step.

\subsection{BYOL}

BYOL \cite{grill2020bootstrap} proposes to bootstrap the representations directly. In particular, BYOL has two networks, referred to as online and target networks. Given two augmented views of the same image, BYOL uses two networks to extract their representations and employs an additional MLP head, called the \emph{prediction head}, to predict the latent representation produced by the target network from the representation produced by the online network. 

\subsubsection{Learn from a randomly initialized model.} One of the core motivations for BYOL \cite{grill2020bootstrap} is an interesting observation: they train a model to predict the representations of a \textit{fixed randomly initialized} model and can reach $18.8\%$ top-1 accuracy in linear probing protocol\footnote{In linear probing protocol, the backbone, \eg, a ResNet, is frozen, while a newly added fully-connected layer is optimized with respect to ImageNet classification \cite{deng2009imagenet}.} on ImageNet, whereas the randomly initialized model itself only achieves $1.4\%$ top-1 accuracy. 

\subsubsection{Improve latents with momentum.} Learning from a randomly initialized model can yield significantly better results (\eg, $+17.4\%$). Therefore, learning from a dynamically improved model seems to be intuitive. In practice, BYOL \cite{grill2020bootstrap} optimizes the online network to predict the target network's latent (feature representation). The target network is updated by a slowly moving exponential average of the online network, \ie, the target network is a momentum copy of the online network. The momentum mechanism is similar to the momentum encoder in MoCo \cite{he2020momentum,chen2020improved}.

\subsection{b-ConCL}

\subsubsection{Bootstrap your own perception.} 
\begin{figure*}[t]
\centering
\includegraphics[width=0.9\linewidth]{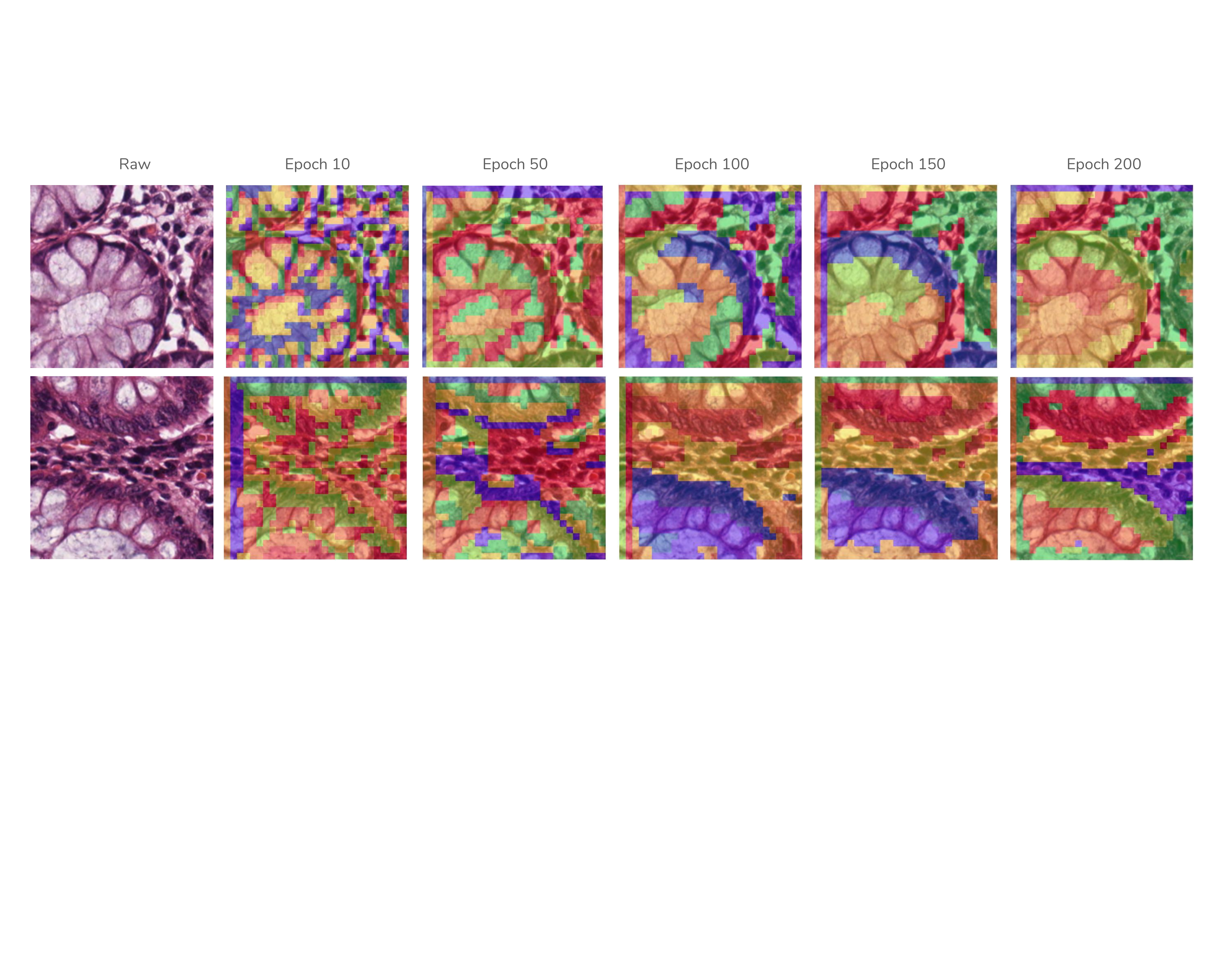}
\caption{\textbf{b-ConCL refines concepts.} We resize images to $448\times 448$ and visualize the concepts clustered from $f_4(\cdot)$ with $K=8$, \ie, 8 clusters. Initially, the grouped concepts are edge-related, but later they become more semantic structure-related.}
\label{fig:bootstrap}
\end{figure*}

b-ConCL resembles BYOL \cite{grill2020bootstrap} in terms of using the cluster results from a momentum key encoder. b-ConCL bootstraps the perception from the momentum encoder while simultaneously improving it and refining it via momentum. \Cref{fig:bootstrap} shows how the concepts grouped by b-ConCL are refined as training continues.

%##################################################################################################
%##################################################################################################

%%%%%%%%%%%%%%%%%%%%%%%%%%%%%%%%%%%%%%%%%%%%%%%%%%%%%%%%%%%%%%%%%%%%%%%%%%%%%%%%%%%%%%%%%%%%%%%%%%%
\section{Additional Implementation Details}\label{sec:add_implementations}

\subsection{ConCL}\label{sec:concl_implementations}

\subsubsection{Data augmentation.} Following MoCo-v2 \cite{chen2020improved}, we use: \texttt{RandomResizedCrop}, and \texttt{ColorJitter} with (brightness=0.4, contrast=0.4, saturation=0.4, hue=0.1) and probability of 0.8, \texttt{RandomGrayscale} with probability of 0.2, \texttt{GaussianBlur} with probability of 0.5, and \texttt{RandomHorizontalFlip}.

\subsubsection{Practical Complement.} In practice, the training process of MoCo \cite{he2020momentum,chen2020improved} is distributed across multiple GPUs. Therefore, features need to be gathered from all GPUs before updating the queue. The gathering operation requires tensors from different GPUs to have the same shape. However, this usually does not hold for ConCL, where the number of the shared concepts (\ie, concepts in both views) varies per pair. We resolve this problem by padding concept keys with randomly re-sampled concepts from the current batch to match the requirement. Specifically, we pad the number of features in each GPU to ${\text{K}*\text{batch\_size}}/{\text{num\_GPUs}}$, where K is the number of clusters. When such padding is infeasible (\eg, K $>$ 8), we only randomly sample 4$*$batch\_size/num\_GPUs concepts. This ensures the validity of the ``gather'' operation. 
    
\subsection{Other Self-supervised Methods}

All self-supervised methods use ResNet-18 \cite{he2016deep} as the backbone. We subsequently modify the numbers of input and hidden channels of the projection head and the prediction head to match the outputs' dimension from ResNet-18. For data augmentation and other hyper-parameters (\eg, temperature, optimizer), we use the default settings specified in each config file in \verb|OpenSelfSup|,\footnote{\url{https://github.com/open-mmlab/OpenSelfSup}} which should be the same as the original methods.

\subsubsection{SimCLR\cite{chen2020simple} \& BYOL \cite{grill2020bootstrap}.} Since both SimCLR and BYOL require a large batch size, we use 1024 for them, which is the maximum number we can afford. For SimCLR, the numbers of channels in each layer of the \textit{projection} head are set to 512-512-128 (input, hidden, output layers). For BYOL, we set them to 512-512-256 (input, hidden, output layers); the numbers of channels in each layer of the \textit{prediction} head are then set to 256-512-256 (input, hidden, output layers).

\subsubsection{MoCo-v1\cite{he2020momentum}, MoCo-v2\cite{chen2020improved} and DenseCL\cite{wang2021dense}.} These three methods require an instance queue. We change the length of the instance queue to 16394 since our pre-training dataset only has 100k samples, far less than 1.28M samples in ImageNet \cite{krizhevsky2012imagenet}. For MoCo-v1, the numbers of channels in each layer of the projection head are set to 512-128 (input, output layers). For the remaining two methods, we set them to 512-512-128 (input, hidden, output layers).  

\subsubsection{PCL-v2\cite{li2020prototypical}.} We use the officially released code\footnote{\url{https://github.com/salesforce/PCL}} for PCL's experiments. We change the numbers of channels in each layer of the projection head to 512-512-128 (input, hidden, output layers) and the number of clusters in PCL to (20000, 35000, 50000).

%%%%%%%%%%%%%%%%%%%%%%%%%%%%%%%%%%%%%%%%%%%%%%%%%%%%%%%%%%%%%%%%%%%%%%%%%%%%%%%%%%%%%%%%%%%%%%%%%%%
\section{Detection Configuration Details}\label{sec:det_config}

\subsection{GlaS.}\label{sec:glas_implementation}

\subsubsection{Introduction.} Glands are important tissue structures for diagnosing adenocarcinomas, a prevalent type of malignant tumor in the prostate, breast, lung, colon, and more. Gland segmentation in pathology images challenge (GlaS) dataset \cite{sirinukunwattana2017gland} collects images from H\&E stained slides with object-instance-level annotation. It consists of a variety of malignant grades. We follow the official train/test split for evaluation, where the training split contains 37 benign and 48 malignant images, and the test split consists of 37 benign and 43 malignant images.

\subsubsection{Transferring setup.} The batch size is 16, and the base learning rate is 0.02. For learning schedules, we refer the $1\times$ fine-tuning schedule to as training for 5k iterations, with the learning rate decayed by ten times smaller at 4k iteration. Similarly, the $0.5\times$ schedule has a total of 2.5k training iterations with decay at 2k iteration; the $2\times$ schedule has a total of 10k iterations, with decay at 8k iteration, and so forth for other schedules.

\subsection{CRAG.}\label{sec:crag_implementation}

\subsubsection{Introduction.} The colorectal adenocarcinoma gland (CRAG) dataset \cite{graham2019mild} collects 213 H\&E stained images taken from 38 WSIs with a pixel resolution of 0.55$\mu$m/pixel at 20$\times$ magnification. Images are mostly of size 1512$\times$1516 with object-instance-level annotation. We follow the official split for evaluation, where the training set has 173 images, and the test set has 40 images with different cancer grades.

\subsubsection{Transferring setup.} The batch size is 16, and the base learning rate is 0.02. For learning schedules, we refer the $1\times$ fine-tuning schedule to as training for 15k iterations, with the learning rate decayed by ten times smaller at 10k and 13k iterations, respectively. Similarly, the $0.5\times$ schedule has a total of 7.5k training iterations with decay at 5k and 6.5k iterations; the $2\times$ schedule has a total of 30k iterations, with decay at 20k and 26k iterations, and so forth for other schedules. The intuition behind the different fine-tuning schedules for GlaS and CRAG is the difference in dataset sizes, \ie, CRAG has around 2.5 times more training samples than GlaS.

\section{More Results}\label{sec:more_results}

\setcounter{table}{0}
\renewcommand{\thetable}{D.\arabic{table}}
\setcounter{figure}{0}
\renewcommand{\thefigure}{D.\arabic{figure}}
    
\subsubsection{Training speed.} Currently, b-ConCL relies on a third reference view for concept matching between different views, which does slow down the training speed. For K-Means, we implement it using matrix multiplication so that the clustering process is parallel to a batch of images. \Cref{tab:training_speed} compares the training speed of different approaches measured in 200-epoch pre-training. Despite our implementation not being optimized thoroughly, ConCL's training speed is satisfactory given the large gains it attains. 

%##################################################################################################
\begin{table}[h]
\centering
\begin{tabular}{c|c|c|c|c|c}
& MoCo-v2 \cite{chen2020improved} & DenseCL\cite{wang2021dense} & PCL\cite{li2020prototypical} & BYOL \cite{grill2020bootstrap} & b-ConCL (ours) \\ \shline
1-epoch	   &	 59.6s	& 65.7s	   & 137.0s & 68.8s & 77.4s  \\ 
\end{tabular}%
\caption{\textbf{Training speeds.} Results are measured in 8-GPU machines. }
\label{tab:training_speed}
\end{table}
%##################################################################################################

\subsubsection{Qualitative comparison of downstream tasks.}

\begin{figure*}[t]
\centering
\includegraphics[width=\linewidth]{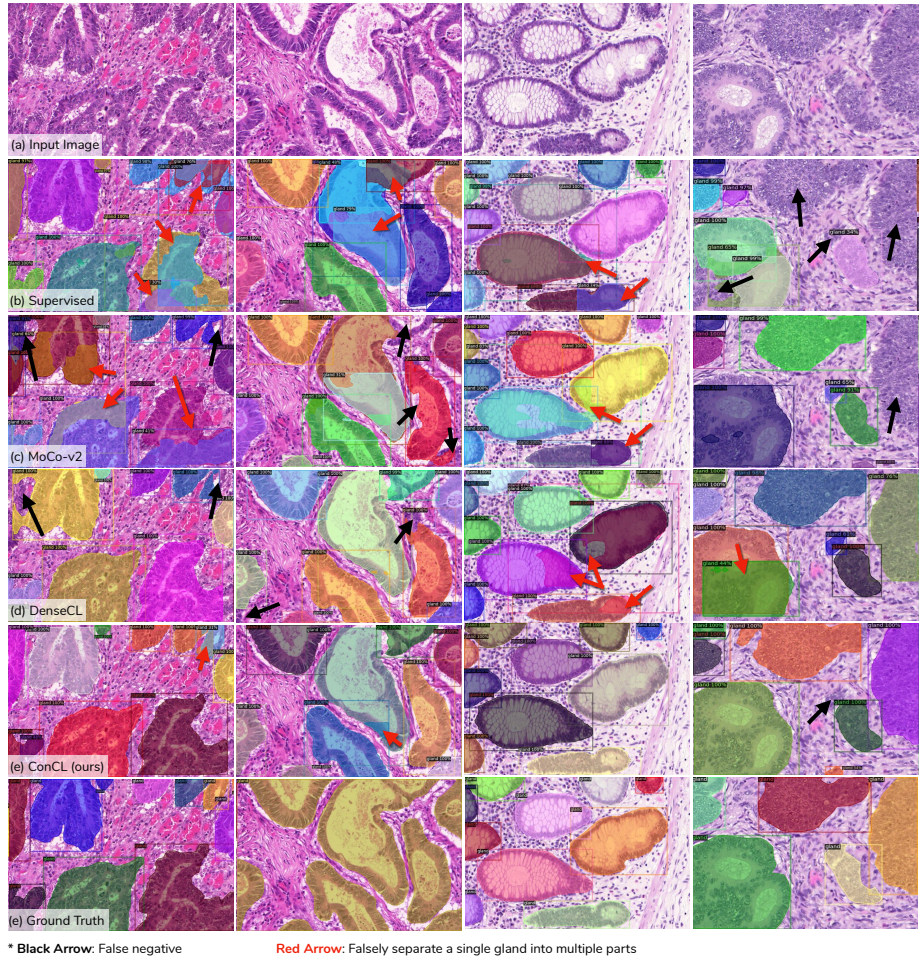}
\caption{\textbf{Qualitative comparison on GlaS dataset \cite{sirinukunwattana2017gland}.} We show the results with Mask-RCNN R18-FPN under $1\times$ schedule.}
\label{fig:glas}
\end{figure*}

\begin{figure*}[t]
\centering
\includegraphics[width=\linewidth]{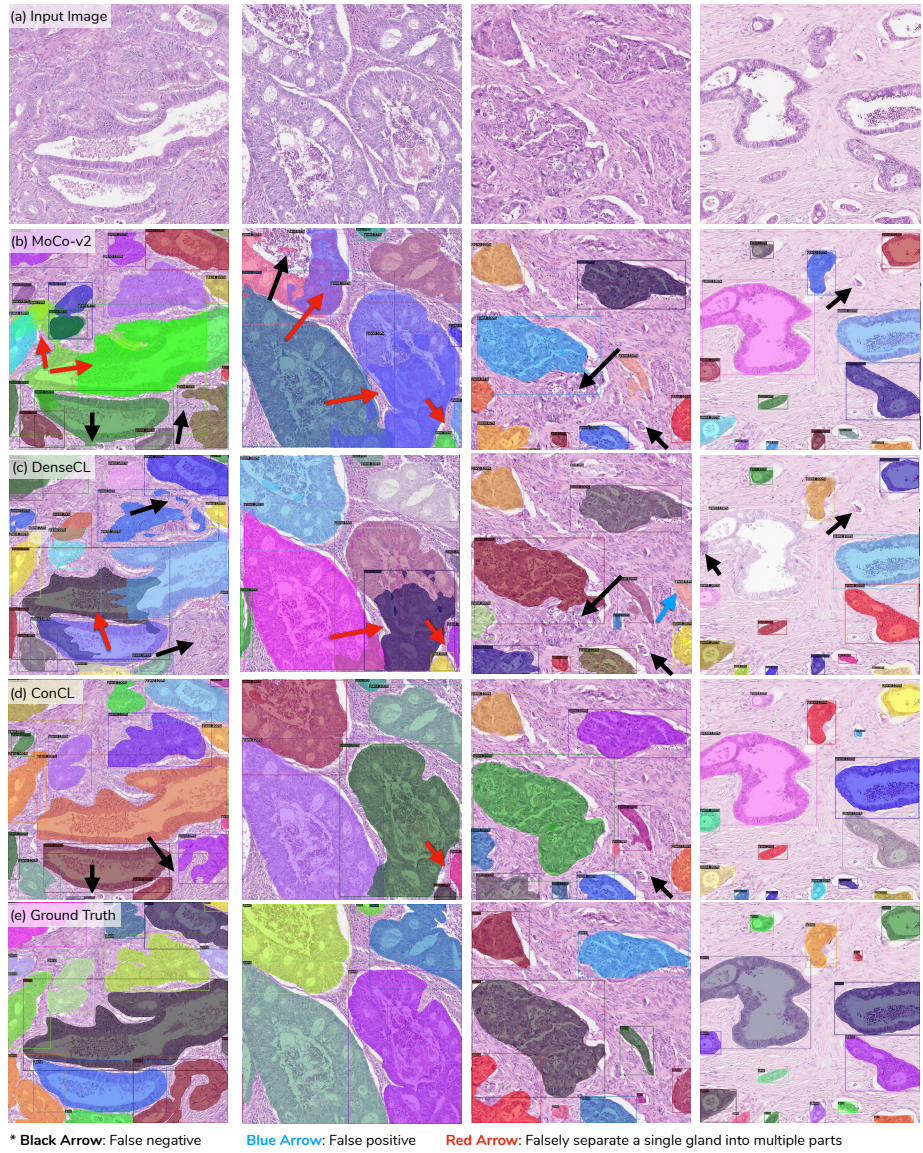}
\caption{\textbf{Qualitative comparison on CRAG dataset \cite{graham2019mild}.} We show the results with Mask-RCNN R18-FPN under $1\times$ schedule.}
\label{fig:crag}
\end{figure*}

\Cref{fig:glas,fig:crag} show some visualizations of different pre-trained models in the GlaS dataset \cite{sirinukunwattana2017gland} and CRAG dataset \cite{graham2019mild}, respectively. We non-exhaustively annotate different detection errors by arrows in different colors. Overall, ConCL outperforms other pre-training methods in terms of less false negative (the black arrows), less false positive (the blue arrows), and more complete detection and segmentation (the red arrows). 

%##################################################################################################
%##################################################################################################

%##################################################################################################
%##################################################################################################

\subsubsection{Numerical results of longer pre-training.}
In complement to \Cref{fig:longer_pretraining1}-(b,c), we further report the numerical results of the transferring performance of 800-epoch pre-trained models in \Cref{tab:longer_pretraining}.

%##################################################################################################
\begin{table}[]
\centering
\begin{tabular}{c|cc|cc}
\multirow{2}{*}{Methods} & \multicolumn{2}{c|}{GlaS} & \multicolumn{2}{c}{CRAG}           \\
                   & AP$^{bb}$ & AP$^{bb}_{75}$ 	& AP$^{bb}$ & AP$^{bb}_{75}$          \\ \shline
SimCLR        &	50.6		     & 56.8		        & 48.1		 	  & 52.0	      \\ 
BYOL          & 50.2             & 56.9             & 49.3            & 54.1              \\
MoCo-v1       & 49.8             & 55.1             & 47.2            & 51.9          \\
MoCo-v2       & 55.2             & 63.6             & 51.8            & 57.6          \\
DenseCL       & \underline{56.0} & \underline{64.8} & \underline{52.5}& \underline{58.2}          \\\hline
b-ConCL (ours)  & \textbf{58.6}    & \textbf{68.1}    & \textbf{55.1}   & \textbf{61.4} \\
\end{tabular}%
\caption{
	\textbf{Transferring performance of different 800-epoch pre-trained models on GlaS and CRAG datasets.}}
\label{tab:longer_pretraining}
\end{table}
%##################################################################################################

\end{document}